\documentclass{article} 
\usepackage{iclr2021_conference,times}

\usepackage{amsmath,amsfonts,bm}

\def\eqref#1{equation~\ref{#1}}

\def\1{\bm{1}}

\DeclareMathAlphabet{\mathsfit}{\encodingdefault}{\sfdefault}{m}{sl}
\SetMathAlphabet{\mathsfit}{bold}{\encodingdefault}{\sfdefault}{bx}{n}

\usepackage{hyperref}
\usepackage{url}
\usepackage{algorithm,algorithmicx,algpseudocode}

\usepackage{amsthm}
\usepackage{booktabs}
\usepackage{xspace}
\usepackage{multirow}

\newcommand{\framework}{PatchGuard++\xspace}

\title{\framework: Efficient Provable Attack Detection against Adversarial Patches}

\iclrfinalcopy

\author{Chong Xiang, Prateek Mittal \\
Princeton University\\
\texttt{cxiang@princeton.edu,pmittal@princeton.edu} \\

}

\begin{document}

\maketitle

\begin{abstract}
An adversarial patch can arbitrarily manipulate image pixels within a restricted region to induce model misclassification. The threat of this localized attack has gained significant attention because the adversary can mount a physically-realizable attack by attaching patches to the victim object. Recent provably robust defenses generally follow the PatchGuard framework~\citep{xiang2020patchguard} by using CNNs with small receptive fields and secure feature aggregation for robust model predictions. In this paper, we extend PatchGuard to \framework for provably detecting the adversarial patch attack to boost both provable robust accuracy and clean accuracy. In \framework, we first use a CNN with small receptive fields for feature extraction so that the number of features corrupted by the adversarial patch is bounded. Next, we apply masks in the feature space and evaluate predictions on all possible masked feature maps. Finally, we extract a pattern from all masked predictions to catch the adversarial patch attack. We evaluate \framework on ImageNette (a 10-class subset of ImageNet), ImageNet, and CIFAR-10 and demonstrate that \framework significantly improves the provable robustness and clean performance. 
\end{abstract}

\section{Introduction}\label{sec-intro}
The adversarial patch attack against computer vision systems allows the adversary to arbitrarily manipulate pixels within a restricted region to induce model misclassification. Compared with the $L_p$ norm bounded adversary, this localized threat model is more realistic in that the adversary can attach the patch to an object to mount a physical attack~\citep{brown2017adversarial,evtimov2017robust}. To counter the threat of an adversarial patch, empirical defenses~\citep{hayes2018visible,naseer2019local} were proposed first but later shown vulnerable to adaptive attackers. The fragility of empirical defenses has inspired provably robust defenses~\citep{chiang2020certified,zhang2020clipped,levine2020randomized,xiang2020patchguard,lin2021certified,metzen2021efficient} for making robust predictions.  

Despite the significant progress in provably robust defenses, however, there is still a huge gap between the provable robust accuracy of defended models and the clean accuracy of undefended models. To narrow this gap, we relax the defense objective from making robust predictions to robustly detecting the attack~\citep{mccoyd2020minority}. This objective is meaningful in applications where ML systems can alert users. For example, if an autonomous vehicle detects an adversarial patch attack, it can alert and return the control to the driver. Towards this end, we propose \framework as an efficient and provably robust attack detection mechanism, which significantly improves both provable robust accuracy and clean accuracy compared with previous provably robust defenses.

In \framework, we first follow PatchGuard~\citep{xiang2020patchguard} to use a CNN with small receptive fields for feature extraction. The small receptive field bounds the number of features that can be corrupted by an adversarial patch. Next, we apply a mask to all possible locations in the feature map and evaluate masked predictions. We empirically observe that the predictions of CNNs are generally invariant to partial feature masking, and thus all masked predictions of clean images usually reach a consensus. On the other hand, when an adversarial patch is attached, some masked predictions are affected by the adversarial perturbations while some are not (when all corrupted features are masked). This will lead to a disagreement in masked predictions. Our final step aims to catch this disagreement for provable attack detection. 
We evaluate our defense on ImageNette~\citep{imagenette} (a 10-class subset of ImageNet), ImageNet~\citep{deng2009imagenet}, and CIFAR-10~\citep{cifar} datasets and demonstrate state-of-the-art defense performance for high-resolution images. Notably, we achieve a 95.3\% clean accuracy and 92.2\% provable robust accuracy on the 10-class ImageNette, against a 2\%-pixel square patch (the patch can be anywhere on the image).

\section{Problem Formulation}\label{sec-formulation}

\textbf{Adversarial Patch Attack.} We focus on the evasion attack against image classification models. Given an image $\mathbf{x} \in \mathcal{X} \subset [0,1]^{W\times H \times 3}$, its true class label $y\in\mathcal{Y} = \{0,1,\cdots,N-1\}$, and an image classification model $\mathcal{M}(\cdot)$, the goal of the attacker is to find an image $\mathbf{x}^\prime \in \mathcal{A}(\mathbf{x})$ satisfying a constraint $\mathcal{A}$ such that $\mathcal{M}(\mathbf{x}^\prime) \neq y$. The constrain $\mathcal{A}(\mathbf{x})$ is determined by the threat model. In this paper, we allow the adversary to arbitrarily manipulate pixels within one square contiguous region and place such restricted region \emph{anywhere} on the image, which is similar to previous works~\citep{levine2020randomized,mccoyd2020minority,xiang2020patchguard}.
Formally, we use a binary \textit{pixel block} $\mathbf{p} \in \{0,1\}^{W\times H}$ to represent the restricted region, where the pixels within the region are set to one. The constraint set $\mathcal{A}(\mathbf{x})$ can be expressed as $\{\mathbf{x}^\prime = (\mathbf{1}-\mathbf{p})\odot \mathbf{x} + \mathbf{p} \odot \mathbf{x}^{\prime\prime} \}$, where $\odot$ refers to the element-wise product operator, and $\mathbf{x}^{\prime\prime}$ is the content of the adversarial patch. 

\textbf{Provably Robust Defense for Attack Detection.} We consider the strongest adaptive attacker who has the perfect knowledge of the defense mechanisms as well as the white-box access to the model architecture and weights. We focus on the \emph{provably robust} defense whose robust guarantee holds for \emph{any} attacker considered in the threat model. The goal of \emph{provable attack detection} is to make normal predictions on clean images but alert when an attack takes place. 


\section{\framework: Provable Attack Detection}

\noindent\textbf{Defense Intuition.} Our key observation is that clean images are generally prediction-invariant to partial feature masking while adversarial patched images are not. When benign features are masked, the model is likely to give the same prediction; when corrupted features are completely masked, the adversary cannot influence the model prediction, and the model prediction is likely to change. Therefore, we can move a mask, which is large enough to mask out all corrupted features, over the feature map, and take the inconsistent masked prediction as an attack indicator. In the remainder of this section, we will introduce our notation, the defense algorithm, and its provable analysis.

\textbf{Notation.} The defense takes an image $\mathbf{x}$ as input and outputs prediction $\mathcal{M}(\mathbf{x})$ or issues an attack alert. The model $\mathcal{M}(\cdot)$ can be any CNN with small receptive fields (relatively small compared with the input image size). We use $\mathcal{F}$ to denote the feature extractor of the model $\mathcal{M}(\cdot)$, and $\mathcal{F}(\mathbf{x})$ gives the feature map $\mathbf{u}\in \mathbb{R}^{W^\prime\times H^\prime \times C^\prime}$. We will abuse the notation to let $\mathcal{M}(\mathbf{u})$ denote making prediction based on the feature map $\mathbf{u}$. We use $\mathbf{w}\in\mathcal{W}\subset \{0,1\}^{W^\prime\times H^\prime}$ to denote a sliding window over the feature space, where elements within the window are set to ones and others are zeros.

\noindent \textbf{Algorithm.} We provide our defense pseudocode in Algorithm~\ref{alg-detection}. We first get the prediction $\bar{y}_0$ and confidence $\bar{c}_0$ from the model with small receptive fields $\mathcal{M}(\mathbf{x})$, as well as feature map $\mathbf{u} = \mathcal{F}(\mathbf{x})$. Next, we try every window mask $\mathbf{w} \in \mathcal{W}$,\footnote{The size of window masks can be computed as $\texttt{w} = \lceil(\texttt{p}+\texttt{r}-1)/\texttt{s}\rceil$, where \texttt{p} is an estimation of the upper bound of patch size, \texttt{r} is the receptive field size, and \texttt{s} is the stride size of receptive field~\citep{xiang2020patchguard}. This window mask size can ensure that at least one window masks out the all corrupted features.} and get masked prediction $\bar{y}_1$ and confidence $\bar{c}_1$ from $\mathcal{M}((\mathbf{1}-\mathbf{w})\odot\mathbf{u})$. If the confidence $\bar{c}_1$ is larger than confidence threshold $\tau$ and the prediction $\bar{y}_1$ is different from $\bar{y}_0$, our defense detects an inconsistent prediction and returns \texttt{alert} (i.e., attack detected). If confidence $\bar{c}_1$ is smaller than $\tau$, we consider this masked prediction ``abstains" and will not affect our final prediction. Finally, if we find all possible non-abstained masked predictions are consistent, we assume there is no attack and return the original model prediction $\mathcal{M}(\mathbf{x})$.

\noindent \textbf{Provable Analysis.} We provide pseudocode for provable analysis in Algorithm~\ref{alg-provable}. Given a clean image $\mathbf{x}$ and its label $y$, we will evaluate all possible masked predictions. If all predictions do not abstain ($\bar{c}_1>\tau$) and are correct ($\bar{y}_1==y$), we will know at least one masked prediction is non-abstained and correct regardless of any attack strategy, including adaptive white-box attacks. Therefore, our defense (Algorithm~\ref{alg-detection}) will either predict $y$ correctly or alert for this certified image.

\begin{minipage}{0.48\textwidth}
\begin{algorithm}[H]
    \centering
    \caption{\framework}\label{alg-detection}
    \begin{algorithmic}[1]
    \renewcommand{\algorithmicrequire}{\textbf{Input:}}
    \renewcommand{\algorithmicensure}{\textbf{Output:}}
    \Require Image $\mathbf{x}$, feature extractor $\mathcal{F}$ of model $\mathcal{M}$, confidence threshold $\tau$, the set of sliding windows $\mathcal{W}$
    \Ensure  Prediction $\mathcal{M}(\mathbf{x})$ or \texttt{alert}
    \Procedure{PG2}{}
    \State $\bar{y}_0,\bar{c}_0 \gets \mathcal{M}(\mathbf{x})$ \Comment{Predict w/o masks}
    \State $\mathbf{u} \gets \mathcal{F}(\mathbf{x})$ \Comment{Extract features}
    \For{each $\mathbf{w} \in \mathcal{W}$}
    \State $\bar{y}_1,\bar{c}_1 \gets \mathcal{M}((\mathbf{1}-\mathbf{w})\odot\mathbf{u})$\Comment{Mask}
    \If{$\bar{c}_1 >\tau$ and $\bar{y}_1 \neq \bar{y}_0$}
    \State\Return \texttt{alert} \Comment{Alert!}
    \EndIf
    \EndFor
    \State\Return $\mathcal{M}(\mathbf{x})$\Comment{No attack}
    \EndProcedure
    \end{algorithmic}
\end{algorithm}
\end{minipage}
\hfill
\begin{minipage}{0.48\textwidth}
\begin{algorithm}[H]
    \centering
    \caption{\framework provable analysis}\label{alg-provable}
    \begin{algorithmic}[1]
    \renewcommand{\algorithmicrequire}{\textbf{Input:}}
    \renewcommand{\algorithmicensure}{\textbf{Output:}}
    \Require Image $\mathbf{x}$, label $y$, feature extractor $\mathcal{F}$ of model $\mathcal{M}$, confidence threshold $\tau$, the set of sliding windows $\mathcal{W}$
    \Ensure  Whether $\mathbf{x}$ has provable robustness
    \item[]
    \Procedure{PG2ProvableAnalysis}{}
    \State $\mathbf{u} \gets \mathcal{F}(\mathbf{x})$ \Comment{Extract features}
    \For{each $\mathbf{w} \in \mathcal{W}$}
    \State $\bar{y}_1,\bar{c}_1 \gets \mathcal{M}((\mathbf{1}-\mathbf{w})\odot\mathbf{u})$\Comment{Mask}
    \If{$\bar{c}_1 <\tau$ or $\bar{y}_1 \neq y$}
    \State\Return \texttt{False} \Comment{Incor./Abst.}
    \EndIf
    \EndFor
    \State\Return \texttt{True}\Comment{Provably robust!}
    \EndProcedure
    \end{algorithmic}
\end{algorithm}
\end{minipage}

\noindent \textbf{Remark.} We note that Minority Report (MR)~\citep{mccoyd2020minority} also adopts a similar idea of masking and prediction consensus. However, MR uses pixel-space masks, which incurs an extremely large overhead (requires expensive model inference for each mask) and makes MR inapplicable to high-resolution images like ImageNet~\citep{deng2009imagenet}. In contrast, the feature-space masking in \framework reduces the number of all possible masks and also reuses the expensive feature extraction computation, making \framework easily scale to ImageNet images. 

\section{Evaluation}\label{sec-eval}
In this section, we present the evaluation results of \framework on different datasets and demonstrate state-of-the-art provable robust accuracy and clean accuracy on high-resolution images. 

\noindent \textbf{Datasets.} We evaluate proposed defenses on three datasets: ImageNet~\citep{deng2009imagenet}, ImageNette~\citep{imagenette}, and CIFAR-10~\citep{cifar}. ImageNet is a 1000-class high-resolution (resized and cropped to 224$\times$224) image classification benchmark; ImageNette is a 10-class subset of ImageNet. CIFAR-10 is a popular image benchmark dataset with a low resolution of 32$\times$32; we rescale it to 192$\times$192 as the input of \framework as done in PatchGuard~\citep{xiang2020patchguard}. 

\noindent \textbf{Setup.} We use BagNet-33~\citep{brendel2019approximating} with a small receptive field of 33$\times$33 as the backbone CNN. Following previous works~\citep{xiang2020patchguard,metzen2021efficient}, we use 2\%-pixel square patch for ImageNette and ImageNet, a 2.4\%-pixel square patch for CIFAR-10.

\textbf{Evaluation.} In Table~\ref{tab-detetcion}, we report the results for \framework and also compare them with previous provably robust defenses. PG-Mask-BN~\citep{xiang2020patchguard}, PG-Mask-BN~\citep{xiang2020patchguard}, and BagCert~\citep{metzen2021efficient} are provably robust for making correct predictions while MR~\citep{mccoyd2020minority} and \framework are provably robust for detecting an attack. For previous defenses, we report results that are available in its original paper. 
For \framework, we report results with different confidence threshold $\tau$.

\textit{\underline{Observation:} \framework works extremely well on high-resolution images, significantly outperforming all previous defenses in terms of provable robust accuracy and clean accuracy.} We report defense performance in Table~\ref{tab-detetcion}. Since MR is computationally infeasible for high-resolution images, we only compare performance with robust prediction defenses on ImageNette and ImageNet. Notably, \framework achieves a high clean accuracy of 96.1\% and a high provable robust accuracy of 91.8\% on ImageNette (setting $\tau=0.6$), outperforming all previous defenses. On ImageNet, \framework has a 6+\% higher clean accuracy, and 13+\% higher provable robust accuracy than all previous defenses when setting $\tau=0.5$! Moreover, if we set $\tau=0.3$ (not listed in the table), we can achieve a similar clean accuracy (56.1\%) as PG-Mask-BN (54.6\%) but a much higher provable robust accuracy (45.4\%; 19.4\% higher!).

\textit{\underline{Observation:} \framework also achieves comparable or better provable robust accuracy on low-resolution images.} The comparison between MR and \framework on CIFAR-10 shows that MR has a slightly better provable robustness in certain settings. However, the computation cost of MR is 100$\times$ higher than ours (making it inapplicable to images in higher resolutions). Moreover, we can see \framework achieves a huge improvement in provable robust accuracy compared with previous robust prediction defenses.

\textit{\underline{Observation:} Threshold $\tau$ balances the trade-off between clean accuracy and provable robust accuracy of \framework.} As shown in Table~\ref{tab-detetcion}, as we decrease $\tau$, the clean accuracy drops while the provable robust accuracy increases (both due to more non-abstained predictions). We note that the provable robust accuracy increases much more quickly than clean accuracy drops, which also demonstrates the strength of our defense design.

\begin{table}[t]
    \centering
        \caption{Clean and provable robust accuracy for different defenses (PG-Mask-BN, PG-Mask-DS, and BagCert are for robust prediction; MR and \framework are for robust detection).}
        \label{tab-detetcion}
      {
     \begin{tabular}{c|c|c|c|c|c|c}
    \toprule
    Dataset & \multicolumn{2}{c|}{ImageNette} & \multicolumn{2}{c|}{ImageNet} & \multicolumn{2}{c}{CIFAR-10}\\

         \midrule
       Accuracy    & clean & robust & clean & robust & clean & robust  \\
       \midrule
       PG-Mask-BN~\citep{xiang2020patchguard} &95.0 & 86.7& 54.6&26.0&83.9&47.3\\
       PG-Mask-DS~\citep{xiang2020patchguard} &92.1&79.7&43.6&15.7&84.6&57.7\\

       BagCert~\citep{metzen2021efficient}& NA & NA& 46.0&22.9&86.0&60.0\\
    \midrule
\multirow{3}{*}{MR~\citep{mccoyd2020minority}}& \multicolumn{4}{c|}{\multirow{3}{*}{computationally infeasible}} & \textbf{92.4}&43.8\\
  &\multicolumn{4}{c|}{}&{90.6}&62.1\\
   &\multicolumn{4}{c|}{}&78.8&\textbf{77.6}\\
   \midrule
   \framework ($\tau=0.8$)  & \textbf{96.9}& 87.7&\textbf{62.9}&28.0 &84.8&68.9\\
    \framework ($\tau=0.7$) & 96.6 & 90.2 & 62.7&32.0 &82.5&71.7\\
    \framework ($\tau=0.6$)  & 96.1 & 91.8 &62.1&35.5 &80.2&74.3\\
    \framework ($\tau=0.5$) & 95.3 & \textbf{92.9} &60.9& \textbf{39.0}& 78.0&76.3 \\
    \bottomrule
    \end{tabular}}
\end{table}

\section{Related Work}

\noindent \textbf{Attacks.} \citeauthor{brown2017adversarial} first proposed a universal targeted adversarial patch attack in the physical world to hijack the model prediction~\citep{brown2017adversarial}. \citeauthor{karmon2018lavan} later proposed a more powerful LaVAN attack in the digital domain~\citep{karmon2018lavan}. More following up works also showed that the localized adversary can achieve a robust physical attack against traffic sign recognition, human detection~\citep{evtimov2017robust,chen2018shapeshifter}. Effective defenses against these realistic attacks are needed.

\noindent \textbf{Defenses.} Heuristic-based Digital Watermarking~\citep{hayes2018visible} and Local Gradient Smoothing~\citep{naseer2019local} defenses were proposed to tackle the threat of localized attacks. However, they had been shown vulnerable to an adaptive attacker~\citep{chiang2020certified}. To provide a stronger security guarantee, \citeauthor{chiang2020certified} proposed the first provable defense against the localized patch attack using Interval Bound Propagation (IBP)~\citep{gowal2018effectiveness,mirman2018differentiable}. Afterwards, Clipped BagNet (CBN)~\citep{zhang2020clipped}, De-randomized Smoothing (DS)~\citep{levine2020randomized}, Minority Report (MR)~\citep{mccoyd2020minority}, PatchGuard~\citep{xiang2020patchguard}, BagCert~\citep{metzen2021efficient} were proposed to improve provable security guarantee against the adversarial patch attack. \framework proposed in this paper can be viewed as a hybrid of PatchGuard (small receptive fields + feature masking) and MR (checking consensus of masked predictions), and it significantly outperforms previous provably robust defenses. 

\section{Conclusion}
In this paper, we propose \framework for provably detecting localized adversarial patch attacks. \framework is an extension of PatchGuard, and achieves provable robustness by detecting inconsistency in all possible feature-masked predictions. Our evaluation on three benchmark datasets demonstrates that our defense significantly boosts clean performance and provable robustness on high-resolution images.

\bibliography{pg2}
\bibliographystyle{iclr2021_conference}


\end{document}